\documentclass[11pt]{article}
\usepackage[pctex32]{graphics}
\usepackage{amssymb}
\usepackage{latexsym}
\usepackage{epsfig} 
\usepackage{multirow} 
\usepackage{pstcol}
\usepackage{float}
\usepackage{epstopdf}
\usepackage{mathrsfs}
\usepackage{soul}

\usepackage{amsmath}
\usepackage{amssymb}
\usepackage{relsize}
\usepackage{multirow}

\usepackage{graphicx}
\usepackage{multicol} 

\usepackage{times}
\usepackage{epstopdf}

\definecolor{Red}{rgb}{1,0.,0.}

\usepackage{fancyhdr}
\usepackage{setspace}

\usepackage{hyperref}
\usepackage{mathtools}
\usepackage{algorithm}
\usepackage[dvipsnames]{xcolor}
\usepackage{algorithmic}
\usepackage{placeins}
\newcommand{\R}{{\mathbb R}}

\newcommand{\mT}{{\mathsf T}}
\newcommand{\mA}{{\mathsf A}}
\newcommand{\mE}{{\mathsf E}}
\newcommand{\mI}{{\mathsf I}}
\newcommand{\mC}{{\mathsf C}}
\newcommand{\mD}{{\mathsf D}}
\newcommand{\mH}{{\mathsf H}}
\newcommand{\mW}{{\mathsf W}}

\newcommand{\mV}{{\mathsf V}}
\newcommand{\mU}{{\mathsf U}}
\newcommand{\mSigma}{{\mathsf \Sigma}}
\newcommand{\mGamma}{{\mathsf \Gamma}}

\title{ Bayesian sparsity and class sparsity priors for dictionary learning and coding}
\author{A Bocchinfuso\footnote{Current address: Oak Ridge National Laboratory} \and D Calvetti \and E Somersalo}
\date{Department of Mathematics, Applied Mathematics, and Statistics\\
	Case Western Reserve University \\
	10900 Euclid Avenue \\
	Cleveland, OH 44106}

\begin{document}
\maketitle

\begin{abstract}
Dictionary learning methods continue to gain popularity for the solution of challenging inverse problems. In the dictionary learning approach, the computational forward model is replaced by a large dictionary of possible outcomes, and the problem is to identify the  dictionary entries that best match the data, akin to traditional query matching in search engines. Sparse coding techniques are used to guarantee that the dictionary matching identifies only few of the dictionary entries, and dictionary compression methods are used to reduce the complexity of the matching problem. In this article, we propose a work flow to facilitate the dictionary matching process. First, the full dictionary is divided into subdictionaries that are separately compressed. The error introduced by the dictionary compression is handled in the Bayesian framework as a modeling error. Furthermore, we propose a new Bayesian data-driven group sparsity coding method to help identify subdictionaries that are not relevant for the dictionary matching. After discarding irrelevant subdictionaries, the dictionary matching is addressed as a deflated problem using sparse coding. The compression and deflation steps can lead to substantial decreases of the   computational complexity. The effectiveness of compensating for the dictionary compression error and using the novel group sparsity promotion to deflate the original dictionary are illustrated by applying the methodology to real world problems, the glitch detection in the LIGO experiment and hyperspectral remote sensing.
\end{abstract}

\section{Introduction}
Model-based interpretation of data is a core task in computational sciences, encompassing various goals like model selection, validation or falsification, estimation of unknown parameters in a model, or solving an inverse problem arising, e.g., in medical imaging.  The increase in computational power, and the current paradigm shift towards data-driven methods have given rise to novel ways of explaining data by relating them to precomputed templates, or previously observed controlled data.
The range of applications where this process arises, referred to as {\em dictionary learning} and {\em dictionary matching},  continues to widen well beyond classical applications to signal analysis \cite{Tosic}, image denoising \cite{Dong,Elad}, medical imaging \cite{Xu,MRF,Medical}, remote sensing \cite{hyperspectral,SAR}, and astrophysics \cite{ligo,ligonoise}. Dictionary learning and dictionary matching are important steps in artificial intelligence engines \cite{Wright,Object}. The logistic problems with pre-trained generative chat robots, which makes use of Natural Language Processing and the ability to organize and process huge collections of language items using the Large Language Models,  include effective and computationally efficient ways to organize large dictionaries as well as the ability to encode new items as a clever combination of dictionary items. In this data intensive era, many dictionaries consist of large collections of items, referred to as atoms. In some cases the dictionary entries are known to belong to different clusters, and often several of the atoms share common features, thus making the dictionary somehow redundant. 

Most dictionary learning applications can be thought of as either a classification or a coding tasks, and in both cases, sparsity may be a desirable feature of the solution. Sparse dictionary coding has a remarkable similarity with searching for a sparse solution of a linear inverse problem, with the unknown of interest being the coefficient vector of the linear combination of the dictionary atoms. In some applications, it is believed a priori that the signals to be encoded can be explained in terms of a few of the dictionary atoms. This belief may be supported by the nature of the signal and by specific choices of the dictionary.  In many signal and image processing applications, a dictionary consisting of  discrete cosine or wavelet transform bases may be employed to obtain a sparse representation \cite{CandesTao, Wright}, in which case the dictionary learning task is akin to a generalized change of representation basis. In other applications, where the dictionary is a collection of atoms which may share some traits with the signal to be explained in terms of the dictionary, sparse coding may be desirable to simplify the interpretation.  A canonical motivational example of sparse dictionary learning is compressed sensing \cite{Donoho}, where high dimensional information can be efficiently recovered and retrieved from few linear observations. Whatever the motivation, algorithms for computing sparse solutions of inverse problems are the engine behind sparse dictionary encoding. 

The literature on sparse solutions of inverse problems is rather extensive.  One popular way to promote sparse solutions is by means of the  basis pursuit \cite{Basis Pursuit}, which penalizes the $\ell_1$-norm of the unknown of interest, a technique referred to as LASSO in the statistics community \cite{Lasso}. Another way to promote sparse solutions is by recasting the inverse problem within a Bayesian framework, and using a hierarchical prior model to express the sparsity belief. It has been shown that for some choices of the hierarchical models, the Maximum A Posteriori solution computed by the Bayesian approach provides a computationally efficient approximation of the $\ell_1$ regularized solution \cite{MEG,L2 Magic,BSC}. A hybrid Bayesian sparsity promoting approach, which eventually switches from a globally convergent sparsity promoting prior to a greedier sparsity promoting model has been recently proposed \cite{Analysis,Pragliola}. 

\section{Outline and novelty}

In this work, we consider the class of dictionary learning/matching problems where the computational complexity is a challenge. Let the matrix $\mD\in\R^{n\times p}$ represent an original dictionary consisting of $p$ atoms in $\R^n$ arranged in its columns. The atoms may be earlier observations of a specific quantity, or they may be synthetic outputs of a computational model. In the case where there is large data set of possible outcomes, we have $p \ll n$, that is, the number of atoms generously exceeds the dimensionality of the data.  In other cases  we may have few atoms of large dimensionality in which case $n>p$. Consider now a new datum $b\in\R^n$ that we want to explain in terms of few of the dictionary atoms. This is equivalent to seeking to solve 
\begin{equation}\label{full}
 b = \mD x + \varepsilon,
\end{equation}
where $x\in\R^p$ is sparse with non-negative entries, and $\varepsilon$ represents the discrepancy between the datum and what is representable in terms of the dictionary. In other words, the coefficient vector $x$ identifies the few atoms needed to explain the data, and the discrepancy is ideally insignificantly small. If the atoms are outputs of a parametrized computational model, such sparse representation can be seen as a way to identify model parameters that best fit to the new observation. Ideally, only one atom is identified, but more realistically, several candidate model outputs may be singled out.

If the the full dataset $\mD$ is the dictionary, one way to reduce the complexity of the problem and simultaneously avoid overfitting is to compute a compressed dictionary by computing a low rank approximating of $\mD$,
\begin{equation}\label{reduce}
 \mD = \mW \mH +\mE, \quad \mW\in\R^{n\times k}, \quad \mH\in\R^{k\times p}, \quad k\ll p.
\end{equation}
It may be desirable to require the entries of the coefficient matrix $\mH$ to be non-negative, making it easier to interpret them as weights. The columns of the matrix $\mW$, referred to as the {\em code book} in  \cite{Aharon}, may be required to retain some properties of the original atoms, e.g., non-negativity or sparsity.  Several algorithms for the model reduction have been proposed, \cite{Aharon}, and often this step is  referred to as dictionary learning in the literature. The matrix $\mE$ accounts for the discrepancy between the original dictionary $\mD$ and its low-rank approximation $\mW \mH$ in terms of the reduced dictionary. After replacing the dictionary with its low rank approximation, expressing the new datum $b$ in terms of the compressed dictionary requires the solution of the linear system
\begin{equation}\label{reduced}
 b = \mW h + \varepsilon',
\end{equation}
where $h\in\R^k$ is non-negative and possibly sparse, and $\varepsilon'$ accounts for the difference between the datum and its approximation in terms of the feature vectors. If the dictionary reduction is performed so that the code book entries are interpretable, the reduced representation may also allow an interpretation for the data $b$. 

In some dictionary applications the atoms are partitioned into disjoint classes and the task is to classify a new datum into one of the classes. A number of algorithms for reducing the computational complexity of dictionary classification have been proposed in the literature, see, e.g., \cite{Metaface,WCS}. The algorithmic approach that we propose here is naturally suitable for dictionary classification also, which may be seen as an intermediate step to reduce the computational complexity of  dictionary matching.
 
\begin{figure}[tbh!]
\centerline{\includegraphics[width=0.5\textwidth]{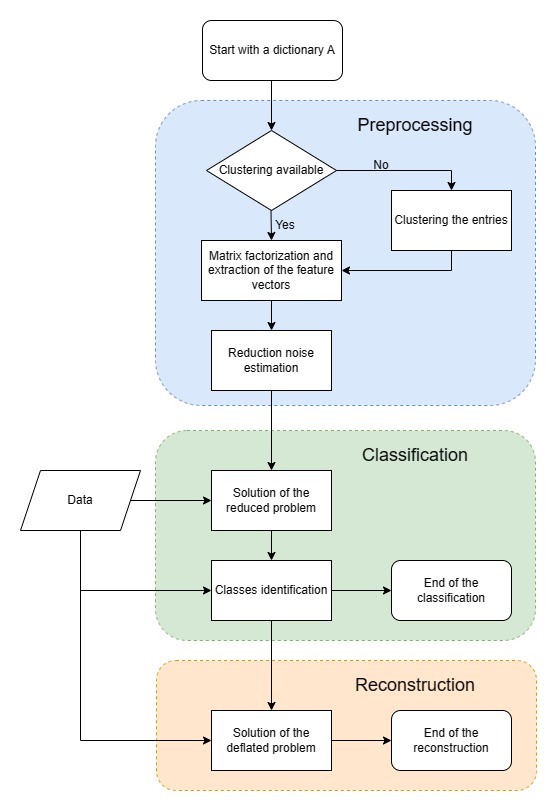}}
\caption{ \label{fig:flowchart} Overview of how the four phases of the proposed algorithm come into play in different application of dictionary learning, matching, or classification.}
\end{figure}

In this article, we propose a multi-phase process of the dictionary learning algorithm, summarized in the following steps.
\begin{enumerate}
\item {\bf Clustering:} Partition the original full dictionary into subdictionaries $\mD^{(j)}$, $1\leq j\leq K$ using supervised or unsupervised clustering.
\item {\bf Reduction:} Compress each subdictionary into a low rank code book, $\mD^{(j)} = \mW^{(j)} \mH^{(j)} + \mE^{(j)}$, $1\leq j\leq K$.
\item {\bf Cluster identification:} Given a new datum $b\in \R^n$, identify reduced subdictionaries that are needed to explain $b$,
\[
 b = \mW^{(j_1)}h^{(j_1)} + \ldots + \mW^{(j_r)}h^{(j_r)} + \varepsilon',
\] 
\item {\bf Deflation:} Represent $b$ in terms of original subdictionaries  $\mD^{(j)}$ that correspond to the  selected compressed dictionaries,
\[
 b = \mD^{(j_1)}x^{(j_1)} + \ldots + \mD^{(j_r)}x^{(j_r)} + \varepsilon.
\] 
\end{enumerate}  
Depending on the objective, one or more of the phases may be skipped. The flowchart in Figure~\ref{fig:flowchart} provides an overview of how the four different phases can be connected.  

The process is carried out in a Bayesian framework. In particular, all steps involving sparse coding are performed by using sparsity promoting Bayesian hypermodels and priorconditioning techniques that have been shown earlier to provide a computationally efficient way to find compressible solutions to linear inverse problems. As a novelty, in the cluster identification problem, we introduce a new, data-informed way to implement group sparsity in order to identify more clearly those clusters that do not contribute to explaining the data. Moreover, our approach borrows ideas from the previous works on Bayesian modeling error analysis to quantify the modeling error $\mE^{(j)}$ when passing from the  full dictionary cluster to the reduced one. We analyze the approximation errors $\varepsilon$ and $\varepsilon'$ using the Bayesian modeling error methods, and demonstrate that in particular the covariance structure of the error induced by the dictionary reduction may have an impact on the quality of the representation. The algorithm is tested with several computed examples.

\section{Dictionary learning/matching: A layered approach}

In this section, we discuss the four steps outlined in the previous section, leaving some of the details for later discussion. The main computational workhorse is a Bayesian hierarchical solver for linear equations with sparsity and possibly positivity constraints. The solver, referred to as the Iterative Alternating Sequential (IAS) algorithm, will be described in detail in a later section.

\subsection{Clustering}

In many applications, the atoms of the dictionary $\mD$ may originate from different sources, or be associated to different models, thus suggesting a natural way of dividing them into subgroups. Even when there are no obvious reasons to partitions the atoms into subgroups, there may be a latent clustered structure in the dictionary which can be highlighted with the help of unsupervised clustering techniques, e.g., $k$-means or $k$-medoids \cite{444}. In the following, we assume that the dictionary atoms can be partitioned into $K$ clusters, and that the dictionary atoms are ordered so that atoms in the same cluster are stored in adjacent columns. This yields a block partitioning of the matrix representation of the dictionary of the form
\[ 
\mD = \begin{bmatrix}
		\begin{array}{c|c|c}
			\mD^{(1)}	&\dots	&\mD^{(K)}
		\end{array}
	\end{bmatrix}, 
\]
where $\mD^{(j)} \in\R^{n\times p_j}$, $p_1+\ldots+p_K = p$.  
We remark that while for some classification problems the annotation of the atoms is provided, in dictionary matching the original annotation of the atoms may not reflect objective similarity between atoms within a class, but may be based on non data-driven criteria. In general, unsupervised clustering may be advisable from the point of view of dividing the dictionary in intrinsically similar batches even if no particular reason for a cluster structure in the data were known. We will not address here how to determine the number $K$ of classes in unsupervised clustering, referring instead to the pertinent literature \cite{444}.

\subsection{Low rank approximation}

A natural way to reduce the complexity of dictionary learning/matching tasks is to replace each subdictionary $\mD^{(j)}$ by a low rank approximation of the form 
\[
\mD^{(j)} \approx \mW^{(j)} \mH^{(j)},\quad \mW^{(j)} \in \R^{n \times k_j}, \; \mH^{(j)} \in \R^{k_j \times p_j}, \quad k_j <p_j.
\]
The columns of $\mW^{(j)}$ are often called the {\em feature vectors} of the subdictionary $\mD^{(j)}$, because the atoms in $\mD^{(j)}$ can be approximated by their linear combinations and often the entries of $\mH^{(j)}$ are all nonnegative.  In some applications it is desirable that the feature vectors retain some of the atoms' characteristics for ease of interpretability. 
Computing a low rank approximation of a matrix is often an intermediate step in a complex task. Among several possible way to compute a low rank approximation of a matrix, the most popular approach, known a principal component approximation (PCA), is based on a truncated singular value decomposition.  Given the singular value decomposition 
\[
\mD^{(j)}= \mU^{(j)} \mSigma^{(j)} \big[\mV^{(j)}\big]^\mT,
\]
of $\mD^{(j)}$, let $\mW^{(j)}$ be the matrix consisting of the first $k_j$ columns $\mU^{(j)}$, and $\mH^{(j)}$ comprise the first $k_j$ rows of $ [\mV^{(j)}]^\mT$ scaled by the first $k_j$ singular values,
\[
\mW^{(j)} = \mU^{(j)}_{k_j} =  \mU^{(j)}(:,1:k_j), \quad \mH^{(j)} =  \Sigma^{(j)}_{k_j} \big[\mV^{(j)}_{k_j}\big]^\mT
=\Sigma^{(j)}(1:k_j,1:k_j)\big[\mV^{(j)}(:,1:k_j)\big]^\mT.
\]
It follows from the well known Eckart-Young-Mirsky theorem that the PCA is the optimal low rank approximation in the sense that it minimizes the Frobenius norm as well as the nuclear norm of the discrepancy; in general,  however, there is no guarantee about the sign of the entries of $\mH^{(j)}$.

When a matrix has all nonnegative entries, it is possible to compute its factorization into the product of two low rank matrices with nonnegative entries, known as nonnegative matrix factorization (NMF) \cite{Gillis}. The nonnegativity of the feature vectors may simplify their interpretation, while the nonnegativity of the weights makes it easier to assess the importance of each feature vector for explaining the columns of the original matrix.  The most popular algorithms for computing a rank $k$ nonnegative matrix factorization of a nonnegative matrix proceed in an iterative fashion, updating one of the factors at a time, while keeping the other fixed. 
More specifically, a non-negative matrix factorization of  $\mD \geq 0$ seeks to determine $\mW \geq 0, \; \mH \geq 0 $ that minimize the approximation error,
		\[
			\|\mD - \mW \mH \|_F. 
			\]		
If $\mW$ is given, the entries of the $\ell$th column of the matrix $\mH$ are the coefficients of the expansions of the $\ell$th column of $\mD$ in terms of feature vectors, stored as columns of $\mW$, thus $\mH$ can be updated column-wise by solving  
		\[
			h^{(\ell)} = {\rm argmin} \| d^{(\ell)} - \mW h \|_2.
		\]
Conversely, given $\mH$, from the observation that 
\[
\|\mD - \mW \mH \|_F = \|\mD^\mT - \mH^\mT \mW^\mT \|_F, 
\]	
it follows that we can update $\mW$ row-wise by similarly solving a least squares problem, see \cite{Gillis,444} for details. 

Replacing each subdictionary of $\mD$ by its low rank approximation yields 
\begin{eqnarray*}
	\begin{bmatrix}
		\begin{array}{c|c|c}
				\mD ^{(1)}	&\dots 	&\mD ^{(K)}
			\end{array}
			\end{bmatrix} &= & 
	\begin{bmatrix}
			\begin{array}{c|c|c}
					\mW ^{(1)} \mH^{(1)} + \mE ^{(1)}	&\dots 	&\mW ^{(k)} \mH^{(K)} + \mE ^{(K)}
				\end{array} 
			\end{bmatrix}\\
			&=&
			\begin{bmatrix}
				\begin{array}{c|c|c}
					\mW ^{(1)}	&\dots 	&\mW ^{(K)}
				\end{array} 
			\end{bmatrix} \begin{bmatrix}			
				\mH^{(1)}	&0 	&\dots 	&0 \\
				0		&\mH^{(2)}	&0	&\vdots \\
				\vdots	&0	&\ddots	&0 \\
				0	&\dots	&0 	 	&\mH^{(K)}
			\end{bmatrix} 	+ \begin{bmatrix}
				\begin{array}{c|c|c}
					\mE ^{(1)}	&\dots 	&\mE ^{(K)}
				\end{array} 
			\end{bmatrix}\\
			&=& \mW \mH + \mE,
\end{eqnarray*}	
where the matrices $\mE^{(j)}$ and $\mE$ represent the approximation errors.
The product $\mW \mH$ is the induced low rank approximation of $\mD$, whose feature vectors are inherited from the feature vectors of the subdictionaries. In the applications of interest in this paper, the matrix $\mW$ can be viewed as a compression of the dictionary $\mD$, thus we will refer to it as {\em{compressed dictionary}}. The number of atoms in the compressed dictionary is the sum of the number of feature vectors of the subdictionaries, which may be chosen to account for the effective dimensionality of the subdictionaries, as will be illustrated in the computed examples section. We remark that, in general, the low rank approximation induced by the partitioning of the dictionary may differ from a low rank approximation of the whole dictionary. The matrix $\mE$ accounts for the error introduced by the reduction of the dimensionality of the dictionary, effectively quantifying how much discrepancy can be expected between the full dictionary and its reduced form. 

When representing a given datum $b$ in terms of the feature vectors constituting the columns of $\mW$,
\begin{equation}\label{approx b}
 b = \mW h + \varepsilon',
\end{equation}
the approximation error $\varepsilon'$ can be thought of as a realization of the random variable with a distribution that can be approximated by considering the sample
\[
 \mE = \mD - \mW \mH.
\]   
Before moving ahead with the algorithm, we consider  the effect of the reduction error through a Bayesian perspective.

\subsubsection{Dictionary compression error}

In the following, let $\mD\in\R^{n\times p}$ denote either the full dictionary, a single subdictionary $\mD^{(j)}$, or, a subset of $i$ subdictionaries combined, $\big[\mD^{(j_1)}, \ldots,\mD^{(j_i)}\big]$. Furthermore, we denote by $\mW$ and $\mH$ the corresponding rank-$k$ approximate matrix factors. The problem considered here is to approximate a given vector $b$ in terms of $\mW$, that is, to find a vector $h$ so that the approximation error in (\ref{approx b}) is as small as possible.

Consider the columns $d^{(\ell)}\in\R^n$ of $\mD$ that allows a perfect representation in terms of the dictionary $\mD$, and denote by $h^{(\ell)}\in \R^k$ the corresponding coefficient vector in the matrix $\mH$.  We have
\begin{eqnarray*}
 d^{(\ell)} = \mD e_\ell  &=&  \mW h^{(\ell)} + \left(\mD e_\ell - \mW h^{(\ell)} \right) \\
 &=&  \mW h^{(\ell)}  + m^{(\ell)}, \quad 1\leq \ell\leq p,
\end{eqnarray*}
where $e_\ell\in\R^p$ is the canonical $\ell$th unit vector, and $m^{(\ell)}$ is the error due to compression, referred to as dictionary compression error (DCE).  Adhering to a probabilistic interpretation of $d^{(\ell)}$ as a realization of a random vector arising from some underlying probability density, the distribution of the compression error can be interpreted as a push-forward of this distribution by the mapping $d^{(\ell)} \mapsto m^{(\ell)}$. Observe that while the vectors $h^{(\ell)}$ may be computed by minimizing the unweighted least squares error between $d^{(\ell)}$ and $\mW h^{(\ell)}$, there is no guarantee that the discrepancies $m^{(\ell)}$ represent scaled white noise.

Having generated the compression error sample $\{m^{(\ell)}\}_{\ell=1}^p$, we compute the mean and covariance,
\begin{eqnarray*}
\mu_{\rm DCE}  & = & \frac{1}{p} \sum_{\ell=1}^{p}  m^{(\ell)} ,  \\
\mC_{\rm DCE} & =  & \frac{1}{p} \sum_{\ell=1}^{p} \big( m^{(\ell)} - \mu_{\rm DCE}\big)\big( m^{(\ell)} - \mu_{\rm DCE}\big)^\mT + \epsilon\, \mI_n,
\end{eqnarray*}
where the term $\epsilon\, \mI_n$, with $\epsilon$ a small positive constant, is added to ensure that the covariance matrix is positive definite.  To take the covariance structure into account in dictionary learning, we whiten the noise $\varepsilon'$ in (\ref{approx b}), by writing 
\[
\mC_{\rm DCE}^{-1/2} (b- \mu_{\rm DCE}) =  \mC_{\rm DCE}^{-1/2} \mW h + w, 
\]
so that  $w$ can be assumed to be mean zero and have covariance close to the identity. This equation gives rise to the {\em dictionary compression error-enhanced least squares problem},
\begin{equation}\label{DCE LSQ}
 h = {\rm argmin}\big\| \mC_{\rm DCE}^{-1/2} (b- \mu_{\rm DCE} - \mW h)\big\|^2_2,
\end{equation}
that can be augmented by appropriate sparsity requirements as specified later.
We point out the similarity of this approach with the techniques previously used to account for the error due,  e.g., to coarse discretization or poorly known model parameters in computational inverse problems, see,  \cite{moderror,moderror2,moderror3}.  

\subsection{Subdictionary identification}

Consider the minimization problem (\ref{DCE LSQ}), assuming that the compressed dictionary $\mW$ consists of all subdictionaries $\mW^{(1)}, \ldots, \mW^{(K)}$. The minimization process yields an approximation of the form
\[
 b \approx \mW^{(1)} h^{(1)} + \ldots + \mW^{(K)} h^{(K)}.
\]
Assume that the datum  $b$ can be explained by few dictionary atoms. Consequently, if the dictionary compression reflects well the structure of the original full dictionary, one could expect that only few of the compressed subdictionaries $\mW^{(j)}$ should be needed to explain the datum $b$ with reasonable accuracy. To find out which subdictionaries are necessary, we adopt a {\em group sparsity} prior for the vectors $h^{(j)}$. Assume that for each $j$, $1\leq j\leq K$, an appropriate norm $\|\,\cdot\,\|_{(j)}$ in $\R^{k_j}$ is defined. The choice of these norms is discussed below. To promote group sparsity, most of the norms $\|h^{(j)}\|_{(j)}$ should vanish or be negligibly small, therefore we want to find $h$ such that
\begin{equation}\label{group sparse}
   h = {\rm argmin}\big\{ \big\| \mC_{\rm DCE}^{-1/2} (b- \mu_{\rm DCE} - \mW h)\big\|^2 +\lambda \big\| \big( \|h^{(1)}\|_{(1)},\ldots, \|h^{(K)}\|_{(K)}\big)\big\|_0\big\},
\end{equation}
where $\|\,\cdot\,\|_0$ indicates the support of the vector and $\lambda>0$. Equivalently, we seek a solution of (\ref{DCE LSQ})  such that the cardinality of the vector of the norms of $h^{(j)}$ is as small as possible. The minimization is performed by using an appropriately defined Bayesian group sparsity promoting algorithm, the details being given later. We discuss next the norms $\|\,\cdot\,\|_{(j)}$, which are constructed by using structural information about the reduced subdictionary representations of the dictionary entries.

\subsubsection{Structural prior for subcluster coefficients}\label{sec:structural}

\begin{figure}[ht]
\centerline{\includegraphics[width=0.5\textwidth]{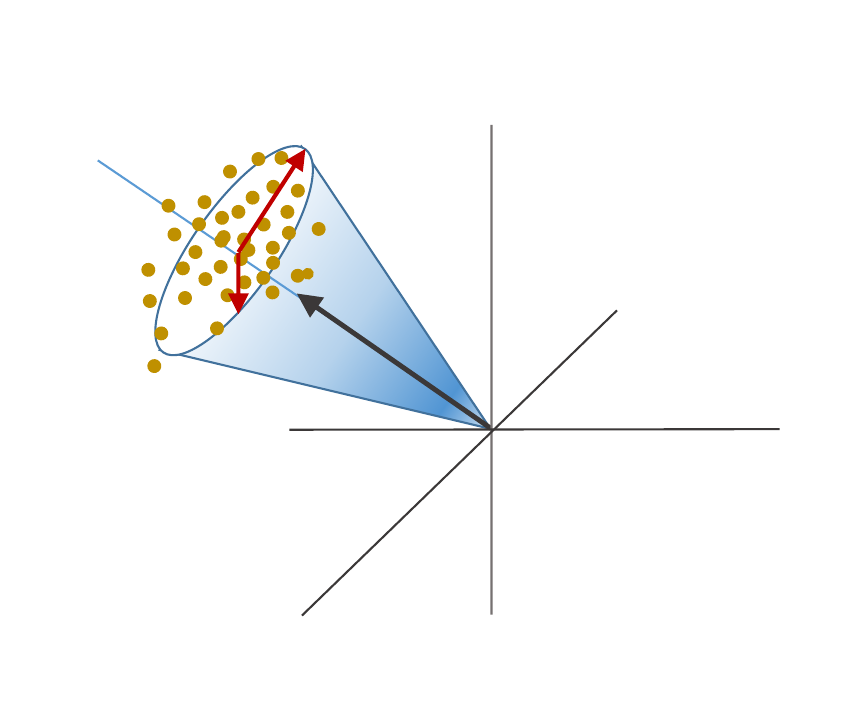}}
\caption{ \label{fig:cone} Schematic picture of the structural prior. The dots indicate the column vectors of the matrix $\mH^{(j)}$. The black arrow is the direction $u^{(j)}_1$ of maximum spread, which in this case is close to the mean of the column vectors, while the red arrows indicate vectors $u^{(j)}_\ell$, $\ell\geq 2$, scaled by the ratios $\sigma^{(j)}_\ell/\sigma^{(j)}_1$. The prior expresses the belief that the vector $h^{(j)}$ has a direction close to $u^{(j)}_1$, and the width and shape of the cone of high probability is determined by the sizes of the singular values relative to the largest one.}
\end{figure}

When solving \ref{DCE LSQ}), we expect  $h^{(j)}$ to be structurally similar to the column vectors of the matrix $\mH^{(j)}$ corresponding to the subdictionary $\mD^{(j)}$.  Observe, however, that we cannot expect $h^{(j)}$ to be close to the group mean of the columns of $\mH^{(j)}$, as we believe a priori that most of the groups $\mW^{(j)}$ contribute little or not at all to explaining $b$. Therefore, we postulate a priori that $h^{(j)}$ should point in the general direction of the columns in $\mH^{(j)}$, with its amplitude controlled by the group sparsity promoting prior. To encode that kind of belief, we consider the singular value decomposition 
\[
 \mH^{(j)}= \mU^{(j)} \mSigma^{(j)} \big[\mV^{(j)}\big]^\mT,  \quad \mSigma^{(j)} = {\rm diag}\big([\sigma^{(j)}_1,\ldots,\sigma^{(j)}_{r_j},0,\ldots,0]\big),
\] 
 and observe that the first columns of $\mU^{(j)}$ identify the directions along which the columns have the largest spread \cite{444}, which is also where we would expect $h^{(j)}$ to have the largest orthogonal projection, regardless of its norm. In light of this observation, we consider the symmetric positive definite matrix 
\begin{eqnarray*}
\mC^{(j)}&=&\left(\frac{1}{\sigma_1^{(j)}}\right)^2 \mU^{(j)} \mSigma^{(j)} \big[\Sigma^{(j)}\big]^\mT \big[\mU^{(j)}\big]^\mT + \epsilon\, \mI \\
&=&  u_1^{(j)} \big[u_1^{(j)}\big]^\mT + \sum_{\ell=2}^{r_j} \left(\frac{\sigma_\ell^{(j)}}{\sigma_1^{(j)}}\right)^2 u_\ell^{(j)} \big[u_\ell^{(j)}\big]^\mT + \epsilon^{(j)}\, \mI,
\end{eqnarray*}
where $r_j$ is the rank of $\mH^{(j)}$ and  $\epsilon^{(j)}>0$ is a small constant ensuring the positive definiteness of the matrix. 
We assume a priori that the {\em direction vector} of unknown $h^{(j)}$, denoted by $\widehat h^{(j)}$ follows the angular central Gaussian measure with covariance $\mC^{(j)}$, defined as the pullback of the Gaussian measure ${\mathcal N}(0,\mC^{(j)})$ with respect to the radial projection $\R^{k_j} \to {\mathbb S}^{k_j-1}$. Alternatively, we may express this prior in terms of  the conditional distribution, given the variance of the length of the vector, as
\[
\pi^{(j)}(h^{(j)}\mid \theta_j ) \propto \exp \left( - \frac{\| h^{(j)} \|^2_{(j)}}{2\theta_j} \right), 
\]
where the structural norm is defined as
\begin{equation}\label{struct norm}
  \| h^{(j)} \|^2_{(j)} = \big[h^{(j)}\big]^\mT\big[\mC^{(j)}\big]^{-1} h^{(j)}.
\end{equation}
This is equivalent to expressing the belief that $h^{(j)}$ has a component along $u_1^{(j)}$ with length regulated by the variance parameter $\theta_j$, and it should remain within the cone defined by the columns of $\mH^{(j)}$ with high probability, see Figure~\ref{fig:cone}. The selection of the parameter $\theta_j$ is part of the group sparsity problem and will be addressed in the sequel.

\subsubsection{Thresholding}

When solving the dictionary representation problem with group sparsity promoting prior, one cannot expect that the norms $\|h^{(j)}\|_{(j)}$ defined by (\ref{struct norm}) for the irrelevant subdictionaries to vanish, but rather we expect some of them to be under a threshold value, similarly to compressibility of a signal in compressed sensing applications. To establish whether a coefficient vector $h^{(j)}$ is significantly large, consider the dictionary compression error covariance $\mC_{\rm DCE}^{(j)}$ computed by using the dictionary entries of the $j$th subdictionary $\mD^{(j)}$ alone. We say that the coefficient vector $h^{(j)}\in\R^{k_j}$ is significant if and only if its contribution to explain the datum $b$ is above the standard deviation of the compression error, that is
\[
 \|\mW^{(j)} h^{(j)}\|^2 > {\rm trace}\big(\mC_{\rm DCE}^{(j)}\big).
\] 
Denote by $J =[j_1,\ldots,j_i\}\subset\{1,2,\ldots,K\}$ the set of the indices of the relevant subdictionaries. 

\subsection{Solving the deflated problem}

The final step in the suggested algorithm is to identify the atoms in the deflated dictionary $\mD^{(J)} = \big[\mD^{(j_1)},\cdots,\mD^{(j_i)}\big]$. To this end, we compute first the dictionary compression error mean and covariance corresponding to the subdictionary $\mD^{(J)}$, denoted by $\mu_{\rm DCE}^{(J)}$ and $\mC_{\rm DCE}^{(J)}$. The final solution is the DCE-enhanced solution, augmented with a sparsity promotion.

\subsection{Bayesian hierarchical linear solvers}							
			
Underdetermined linear systems arise in many important inverse problems applications, requiring to address how to handle contributions from the null space. However, the presence of a significant null space adds flexibility to the solution, making it possible to add additional constraints and objectives such as sparsity and positivity. In this section, we briefly review a sparsity promoting Bayesian hierarchical model that will be used to solve the subproblems 2, 3 and 4 of the proposed layered approach. For further details, we refer to \cite{BSC}.

Consider the linear system of equations 
\begin{equation}\label{b=Ax}
b = \mA x + e , \quad \mA \in\R^{n\times p}
\end{equation}
where the  additive error $e$ is assumed to be zero mean Gaussian distribution with covariance matrix $\mGamma$, thus yielding the likelihood
\[
\pi_{B\mid X}(b\mid x) \propto {\rm exp}\left(-\frac 12(b-\mA x)^\mT \mGamma^{-1}(b - \mA x)\right).
\]
We encode our a priori belief that the solution $x$ will have only very few nonzero entries, or entries above a small threshold value, by modeling $x$ as zero mean Gaussian random variable $X$ with a diagonal covariance matrix,
 \[
X \sim {\mathcal N}(0, \mD_\theta), \; \mD_\theta= {\rm diag}( \theta_1, \ldots \theta_p), \; \theta_j>0.
\]
In the hierarchical approach, we assume further that the variance vector $\theta$ is unknown, and model it as a random variable $\Theta$. Hence, the conditional prior for $X$ is 
\[
	\pi_{X\mid\Theta} (x\mid\theta) \propto 
	\frac{1}{
	\sqrt{{\rm det}(\mD_\theta)}
	} \exp \left( -\frac{1}{2} \Vert \mD_\theta^{-\frac{1}{2}} x \Vert^2  \right)= 
	\exp \left(- \frac{1}{2} \sum_{j=1}^{p} \log  \theta_j  -\frac{1}{2} \Vert \mD_\theta^{-\frac{1}{2}} x \Vert^2  \right).
\]
To model the variance vector $\theta$, observe first that the smaller the variance, the more likely it is that the variable $X$ takes on a value close to zero, therefore if the support of the solution were known, it would suffice to make the corresponding diagonal entries large, while setting all others very close to zero. In line with the Bayesian philosophy, since this information is not known a priori, the variances are modeled as random variables. Furthermore, the variances should be independent of each other, positive and, to promote sparsity, most of them should be very small while a few of them could be large. To achieve such outcome, we assume that the variables $\theta_j$ are independent and distributed according to a fat-tailed distribution such as the gamma distribution,

\[
	\pi_{ \Theta} (\theta \mid \beta, \vartheta) =  \prod_{j=1}^p \frac 1{\Gamma(\beta)\vartheta_j^{\beta}} \theta_j^{\beta -1}  \exp \left( -  \frac{\theta_j}{\vartheta_j}\right),
\]
where the value of the shape parameter $\beta$ is related to the level of sparsity promotion and the values of the scale parameters $\vartheta_j$ are related the sensitivity of $b$ to the entries of $x$; see \cite{BSC} for details. Without loss of generality,  we may assume that the additive error in the right hand side of (\ref{b=Ax}) is zero mean white Gaussian. 
Combing the likelihood and the joint prior according to Bayes' formula yields the posterior density for the pair $(X,\Theta)$ 
\[
\pi_{X, \Theta \mid B} (x, \theta \mid b) \propto    \exp \left( -\frac{1}{2} \Vert b - \mA x \Vert^2 - \frac{1}{2} \Vert D_\theta^{-\frac{1}{2}} x \Vert^2 + 
	 \big( \beta - \frac{3}{2} \big) \sum_{j=1}^{p} \log  \theta_j  - \sum_{j=1}^{n} \frac{\theta_j}{\vartheta_j}  \right) \, ,
\]
which in the Bayesian framework is the solution of the inverse problem. While getting insight in the properties of the posterior distribution requires methods like Markov chain Monte Carlo (MCMC) sampling \cite{pCN}, we restrict ourselves here to numerically approximate  the Maximum a Posterior (MAP) estimate, which is also the minimizer of the Gibbs energy,
\[
	{\mathscr E}(x, \theta) = \frac{1}{2} \Vert b - \mA x \Vert^2 + \frac{1}{2} \sum_{j=1}^{p} \frac{x_j^2}{\theta_j} - \big( \beta - \frac{3}{2} \big)\sum_{j=1}^{p} \log  \theta_j + \sum_{j=1}^{p}  \frac{\theta_j}{\vartheta_j}  \, .
\]
A computationally efficient way of approximating the MAP estimate, in particular for large matrices $\mA$, is provided by  the Iterative Alternative Sequential (IAS) algorithms, discussed in detail in \cite{BSC}. The algorithm proceeds by updating sequentially $x$ and $\theta$, taking advantage of the fact that the update of $x$ amounts to the solution of a linear least squares problem, and the update of $\theta_1, \ldots, \theta_p$ can be done independently via a simple formula evaluation. Moreover, the computationally efficient version of the IAS uses the technique of priorconditioning together with Krylov subspace iterative schemes with early stopping. The outline of the algorithm is summarized below, with the details omitted. We point out that the algorithm can be made even more efficient by combining gamma distributed hyperpriors with generalized gamma distributions in a hybrid manner, see \cite{Hybrid,Analysis,pCN}. The hybrid algorithm will be used in the last of the computed examples to enhance the sparsity promotion. Rather than reviewing details that are not central for this article, we refer interested readers to the cited articles.

\bigskip
\hrule

{\bf IAS algorithm}
\medskip

\hrule
\begin{itemize}
\item[] {\bf Given:} matrix $\mA\in\R^{n\times p}$, $b\in\R^n$, $\beta$, $\vartheta_j, \; 1 \leq j \leq p$,
\item[] {\bf Initialize:}  $\theta^{(0)}_j = \vartheta_j, \; 1 \leq j \leq p$, set the counter $k=0$.
\item[] {\bf Iterate} until stopping criterion is met:
\begin{enumerate}
\item Update $x$:  Set $\theta_j = \theta^{(k)}_j, \; 1\leq j \leq p$ and set 
\[
x^{(k+1)} = {\rm arg}\min \big\{ \| b - \mA x\|_2^2 + \|\mD_{\theta^{(k)}}^{-\frac12} x \|_2 ^2 \big\}.
\]
\item Update $\theta_j, \; 1 \leq j \leq p\;$by solving
\[
 \frac{\partial {\mathscr E}}{\partial\theta_j}(x^{(k+1)}, \theta) = 0,
\]
yielding 
\[
\theta_j^{(k+1)} =  \frac{\vartheta_j}{2}\left( \eta + \sqrt{\eta^2 + \frac{[x_j^{(k+1)}]^2}{4\vartheta_j}}\right), \quad \eta = \beta - \frac 32.
\]
\item Advance the counter $k\to k+1$ and check convergence.
\end{enumerate}

\bigskip
\hrule
\bigskip
\end{itemize}

The properties of the IAS algorithm, and its generalizations using generalized gamma distributions have been reported in the literature. When using the gamma distribution for the variances, the functional minimized by the IAS algorithm is convex with a unique global minimum, and the IAS iterates converge to the unique minimizer. It has been shown that, in the limit as $\eta = \beta - 3/2 \rightarrow 0^+$, the solution computed with the IAS algorithm converges to a scaled $\ell_1$-regularized solution, 
\[
 x_{\ell_1} = {\rm argmin} \left\{ \frac 12 \|b - \mA x\|^2 + \sqrt{2} \sum_{j=1}^p \frac{|x_j|}{\sqrt{\vartheta_j}}\right\},
\]

suggesting that the selection of the value of the parameter $\beta$  depends on how sparse the solution is believed to be. In general, the value of $\eta =\beta-\frac32>0$ should be very close to zero if the solution $x$ is believed to be very sparse, while with larger values of $\eta$ the solution computed by the IAS will be closer to an $\ell_2$-regularized solution. 

Moreover, the above formula indicates that the variables $\vartheta_j$ provide a component-wise scaling, and it was demonstrated that 
under the hypothesis that all components $x_j$ must have the same chance to explain the data within a given signal-to-noise ratio, the 
proper choice of the scaling parameters is related to the sensitivity of the data on the components $x_j$, see \cite{L2 Magic,BSC} for details. Moreover, if the entries of the solution are subject to bound constraints, e.g., $x_j\geq 0$, the IAS algorithm can be modified so as the ensure that the constraints are respected by concatenating the IAS iterations with a projection onto the feasible set to the update of $x$, see \cite{Pragliola}. 
 
In the present dictionary learning context, we employ the IAS algorithm in the  NNLS-NMF algorithm to update the columns  of $\mH$  and rows of $\mW$, with the projection on the nonnegative cone guaranteeing the nonnegativity of the entries of $\mH$, see \cite{WCS} for details. If each dictionary entry is to be explained in terms of few features vectors, a very small value of $\eta>0$ is used, while a larger values of $\eta$ is used for the computation of feature vectors that are not expected to be sparse. Observe that the positivity constraint for $\mW$  can also be applied if the feature vectors are required to be non-negative to allow interpretation in terms of the dictionary. Furthermore, the IAS algorithm will be employed also in the phase of solving the deflated problem with $\mA = \mD^{(J)}$, the group reduced full dictionary, where the sparsity of the coefficient vector $h$ is crucial. 

\subsection{IAS with group sparsity}

We consider now the step of identifying relevant clusters by using the reduced subdictionaries, and in particular, the minimization problem (\ref{group sparse}) by using the IAS algorithm.

The goal here is to find an approximate representation
\[
b = \mW h +\varepsilon', 
\]
where 
\[
\mW =  \left[ \begin{array}{c|c|c} \mW^{(1)} &  \ldots & \mW^{(K)} \end{array}\right],  \quad 
h = \left[ \begin{array}{c} h^{(1)} \\  \vdots \\ h^{(K)} \end{array} \right], \quad h^{(j)} \in \R^{p_j},
\]
with the prior belief that most of the vectors $h^{(j)}$ are insignificant.

Define a group sparsity promoting structural prior for $h$, 
\begin{eqnarray*}
 \pi_{H\mid\Theta}(h\mid\theta) &\propto& \prod_{j=1}^K
 \frac1{\sqrt{{\rm det}(\theta_j \mC^{(j)})}}\ 
 {\rm exp}\left( - \sum_{j=1}^K \frac{[h^{(j)}]^\mT [\mC^{(j)}]^{-1} h^{(j)}}{2\theta_j}\right) \\
 & \propto & \exp\left( -\frac12 \sum_{j=1}^K {p_j} \log\theta_j  - \frac12 \sum_{j=1}^K \frac{\|h^{(j)}\|^2_{(j)}}{\theta_j} \right)\\
& \propto & \exp\left( -\frac12 \sum_{j=1}^K {p_j} \log\theta_j  - \frac12 \| \mD_\theta^{-1/2} h \|_2^2 \right),
\end{eqnarray*}
where now $\mD_\theta$ is the block diagonal matrix 
\[
\mD_\theta = \left[ \begin{array}{c c c c } \theta_1 \mC_1 & 0 & \ldots & 0 \\ 0 & \theta_2 \mC_2 & \ddots & \vdots \\ \vdots & \ddots & \ddots & 0 \\ 0 & \ldots & 0 & \theta_K \mC_K \end{array} \right]
\]
and $\mC^{(j)}\in\R^{p_j\times p_j}$ are the structural covariance matrices constructed on the basis of the cone condition of subsection~\ref{sec:structural}.

To promote group sparsity, most $\theta_j$s should be close to zero, with a few large outliers. As in the basic form of the IAS algorithm, we achieve this by postulating that the variables $\theta_j$ are independent and distributed according to a fat-tailed distribution such as the gamma distribution. Combining the likelihood, enhanced by the dictionary compression error, we find that the posterior density in this case assumes the form
\[
\pi_{H,\Theta\mid B}(h, \theta\mid b) \propto \exp \left( -\frac 12 \| \mC_{\rm DCE}^{-12}(b - \mu_{\rm DCE} - \mW h)\|^2_2 - \frac 12\sum_{j=1}^K \frac {\|h^{(j)}\|^2_{(j)}}{\theta_j} + 
2 \sum_{j=1}^K \eta_j \log\theta_j - \sum_{j=1}^K\frac{\theta_j}{\vartheta_j}\right),
\]
where
\[
  \eta_j =  \beta_j- \frac{p_j+2}{2}, \quad 1\leq j\leq K.
\]
The MAP estimate corresponding to this posterior density can be computed by the Group Sparsity IAS (GS-IAS), a version of the IAS algorithm that promotes group sparsity. As the standard IAS, the IAS algorithm for group sparsity proceeds by a sequential iteration that alternates between the update of $h$ with fixed $\theta$, and the update of $\theta$ with fixed $h$.  

\bigskip
\hrule

{\bf GS-IAS algorithm}
\medskip

\hrule
\begin{itemize}
\item[] {\bf Given:} matrix $\mW$, $b$, $\beta_j$, $\vartheta_j, \; 1 \leq j \leq K$,
\item[] {\bf Initialize:}  $\theta^{(0)}_j = \vartheta_j, \; 1 \leq j \leq K$, set the counter $k=0$.
\item[] {\bf Iterate} until stopping criterion is met:
\begin{enumerate}
\item Update $h$:  Fix $\theta_j = \theta^{(0)}_j, \; 1\leq j \leq K$ and set 
\[
h^{(k+1)} = {\rm arg}\min \big\{  \| \mC_{\rm DCE}^{-12}(b - \mu_{\rm DCE} - \mW h)\|^2_2 + \|\mD_{\theta}^{-\frac12} h \|_2 ^2 \big\}.
\]
\item Update $\theta_j, \; 1 \leq j \leq K\;$ by setting 
\[
\theta_j^{(k+1)} =\frac{\vartheta_j}{2}\left(\eta_j  + \sqrt{{\eta_j^2} + \frac{\|h^{(j)}\|_{(j)}^2}{4 \vartheta_j} }\right),
\]
where 
\[ \quad \eta_j = \beta_j - \frac{p_j+2}{2}.
\]
\item Update the counter $k\to k+1$ and check the convergence.
\end{enumerate}

\bigskip
\hrule
\bigskip

\end{itemize}

\section{Computed experiments}

We demonstrate the viability of the novel ideas of this article with three computed examples. The first one is a toy problem using the MNIST handwritten digits, demonstrating the performance of the full proposed algorithm from dictionary clustering to dictionary matching with the deflated dictionary. The second example addresses a more realistic problem discussed previously in the literature, related to the identification of glitches in the LIGO/VIRGO experiment for detecting gravitational waves.  The third example is a remote sensing application based on hyperspectral imaging and identification of the terrain characteristics. The last two examples highlight the importance of the central new features of this article, the dictionary compression error and the structural prior in group sparsity pursuit. 

\subsection{Sparse dictionary coding  applied to handwritten digits}

\begin{figure}[t]
	\centerline{
	\includegraphics[width=0.9\linewidth]{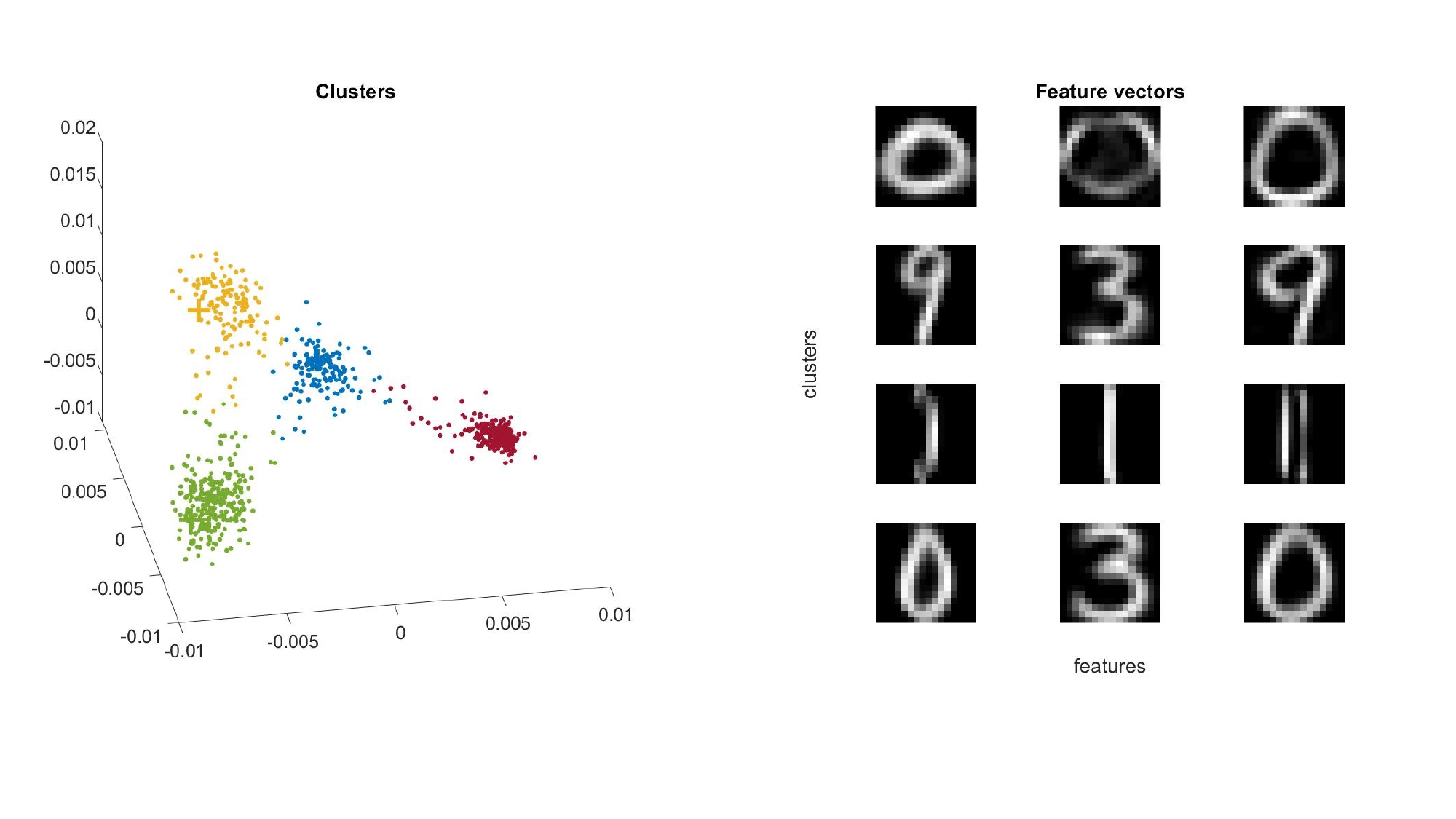}
	}
	\caption{Left: Scatter plot of the four clusters $\mD^{(j)}$, visualized by projecting the data on the first three (non-orthogonal) direction vectors obtained by linear discriminant analysis  (LDA). The unsupervised clustering does not fully correspond to the classification corresponding to the annotation. Right: The three feature vectors in the reduced dictionary $\mW$ for each of the four clusters, found by the $k$-medoids algorithm. Observe that the clustering is not following the annotation, as some of the clusters contain mixtures of features belonging to different annotated groups}
	\label{fig:digits_clusters}
\end{figure}

To demonstrate the main features of the proposed algorithm, we consider first a simple example that allows easily interpretable visualizations. Starting with the standard MNIST database of $16\times 16$ pixel images of handwritten digits stored as vectors in $\R^{256}$, we generate a dictionary matrix  $\mD$ consisting of $p =834$ samples of the four digits  0, 1, 3, and 9, hence $\mD \in \R^{256\times 834}$.  We then run the $k$-medoids algorithm with an $\ell^2$-distance matrix (see \cite{444})  to cluster the data in four clusters. Observe that the clustering is done in an unsupervised manner, that is, the known annotation is not used to guide the clustering.  The clustered vectors are then arranged in four subdictionaries, $\mD^{(1)},\ldots, \mD^{(4)}$. The clustering result is visualized on the left in Figure \ref{fig:digits_clusters} by plotting the three LDA components \cite{444}.

The images are non-negative, and we use a non-negative matrix factorization, computed by using the NNLS-IAS algorithm 
\cite{WCS} with non-negativity constraint to extract three feature vectors per each cluster.  The  feature vectors for each class are shown on the right in Figure~\ref{fig:digits_clusters}, plotted as black-and-white images. The figure indicates that the clusters do not correspond to the annotated classes, as the feature vectors of some clusters do not represent the same digit.  For the present algorithm, this is not a problem since correct clustering in terms of the annotation information is of no consequence here.

The test data consist of $4\,000$ vectors, each one being linear combinations of one, two, or three digits drawn randomly from different subdictionaries of $\mD$ with positive  coefficients.  In particular, the data is within-sample, i.e., drawn from the data set constituting the dictionary.  The performance of the algorithm is measured by its capability of correctly identifying the classes constituting each data vector. To define a quantifier for the success, let $x\in\R^{834}$ denote the coefficient vector defining the data vector, $b = \mD x$, and let  $\widehat x\in\R^{834}$ be the coefficient vector found by the algorithm. We define the thresholded supports of the vectors,
\[
 S_\delta = \{j \mid x_j>\delta\}, \quad  \widehat S_\delta = \{j \mid \widehat x_j>\delta\}, \quad \delta>0,
\]
and define the dissimilarity index as the cardinality of the symmetric difference of them,
\[
 I_\delta (x,\widehat x) = {\rm card}(S_\delta \triangle \widehat S_\delta) = {\rm card}((S_\delta \cup S_\delta)\setminus (S_\delta \cap S_\delta)\ ) .
 \] 
When $I_\delta(x,\widehat x) = 0$, the algorithm has correctly identified the atoms that constitute the data.  Table~\ref{tab:errors} shows that in over 95\% of cases, the dissimilarity index vanishes, indicating that the algorithm is capable of identifying perfectly the constituting atoms.

\begin{figure}
\centerline{\includegraphics[width=0.6\linewidth]{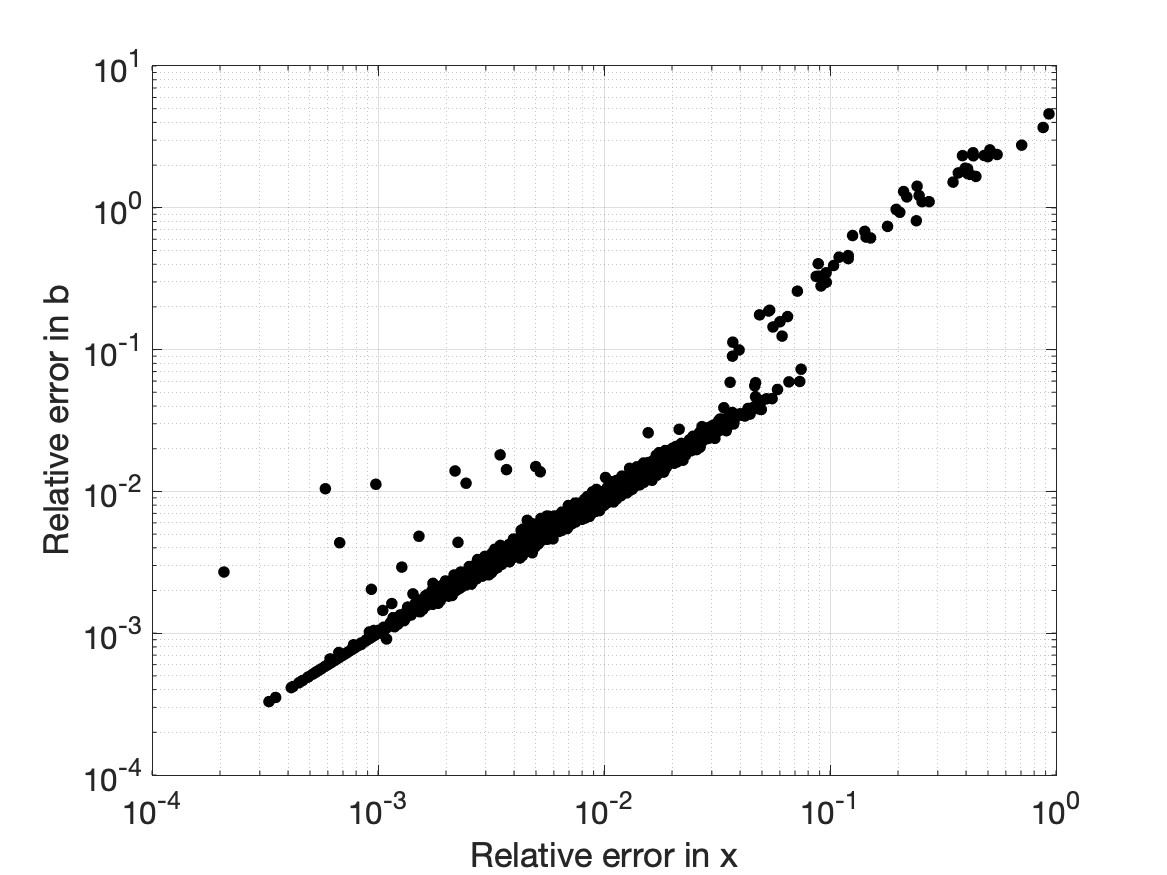}}
\caption{\label{fig:rel error}Relative errors in the reconstruction versus relative error in the data.}
	\label{fig:relative_error}
\end{figure}

The failure of the algorithm can be defined in terms of the relative error in the reconstruction of the coefficient vector,
\begin{equation}\label{rel error x}
r_x(x,\widehat x)  = \frac{\|x-\widehat x\|_2}{\|x\|_2},
\end{equation}
a quantity that cannot be computed without the knowledge of the true $x$. However, it turns out that the relative error in $b$, defined as
 \begin{equation}\label{rel error b}
 r_b(x,\widehat x) = \frac{\|b - \mD \widehat x\|_2}{\|b\|_2} = \frac{\|\mD(x-\widehat x)\|_2}{\|\mD x\|_2},
\end{equation}
which is a quantity that can be computed directly from the data, provides a reliable proxy for (\ref{rel error x}). Figure~\ref{fig:rel error} shows the scatter plot of the two relative errors in logarithmic scale, revealing a strong correlation between the two quantities. Consequently, serious failures of the algorithm can be directly detected by thresholding the relative error in the data

\begin{table}
\centerline{
\begin{tabular}{l|cccccc}
frequency & 3904 (97.6\%) & 34 (0.85\%) &1 (0.02\%) &1 (0.02\%) &1 (0.02\%) &59 (1.48\%) \\
\hline
 $I_\delta(x,\widehat x)$ & 0 & 1 & 2 & 3 & 4 & $\geq 5$ 
\end{tabular}
} 
\caption{\label{tab:errors}The frequencies of occurrency of dissimilarity index values. The total sample size is 4000, and $I_\delta(x,\widehat x)=0$ means perfect identification of the atoms, while $I_\delta(x,\widehat x) = k>0$ indicates that $k$ atoms were either not identified or an incorrect identification occurred.}
\end{table}

Finally, we consider the effect of the dictionary compression error (DCE), which is one of the main novelties of this article. We consider a case in which the data corresponds to a single atom out of a single class, and we perturb the data by Gaussian white noise with standard deviation 0.03. We run the algorithm using 5000 simulations, with and without the inclusion of the DCE. Including the DCE leads to a correct cluster identification in 4835 cases, or 97\% of the cases. Leaving out the DCE, the correct class is identified 4705 times, or 0.94\% of the cases. Not surprisingly, once the correct class is identified, the correct atom is identified in both simulations in approximately 98\% of the cases. The effect of the DCE is observable in this case, but not as crucial as in the following example.

\subsection{Dictionary classification for gravitational wave glitch identification}

In the second example, we consider the dictionary learning and matching problem related to the LIGO/VIRGO experiment in which the observed gravitational wave, or a coincident event of any unknown origin registered by the devices, is to be identified based on a precomputed dictionary, as finding a universal predictive parametric model is not feasible.  
 Due to the extremely high sensitivity of the interferometer, the data stream is contaminated by frequent artifacts of non-gravitational origin, known as glitches, that due to their high occurrence rate may lead to false coincidence detections. The glitches are transients with particular morphologies that allow them to be classified and identified by using precomputed glitch libraries. The examples considered in this work follow the simplified glitch morphology described in \cite{Llorens-Montequado}. In the referred article, the authors describe three morphologically different glitch classes, sine Gaussian (SG), Gaussian (G), and ring-down (RD)  signals,
parametrized as
\begin{eqnarray*}
 h_{\rm SG}(t) &=& h_0 \sin \big(2\pi f_0(t-t_0)\big) e^{-(t-t_0)^2/2\tau^2}, \\
 h_{\rm G}(t)  &=& h_0 e^{(t-t_0)^2/2\tau^2}, \\
 h_{\rm RD}(t) &=& h_0\sin \big(2\pi f_0 (t-t_0)\big) e^{-(t-t_0)/\tau}\theta(t-t_0), 
 \end{eqnarray*}
 where $h_0$ is the amplitude, $f_0$ is the frequency, $t_0$ is the center time and $\tau$ is the characteristic time of the glitch, and $\theta$ is the Heaviside function.  The morphology  of real LIGO glitches is wider than this, see, e.g., \cite{Powell1, Powell2} for typical spectrograms.  

  We begin by generating a model-based library of signals. The duration of each signal is one second, and the sampling frequency is 16\,384 Hz as in the LIGO experiment, leading to a dictionary with $m = 16\,384$.  We generate 1\,000 atoms per class. The model parameters are drawn randomly as in  \cite{Llorens-Montequado}: For SG and RD waveforms, $f_0$ (Hz) is drawn from log-uniform distribution over the interval $[40,1500]$ Hertz, and consequently,
 $\tau$ is set as $\tau = Q/\sqrt{2} \pi f_0$, where the quality factor $Q$ is drawn from log-uniform distribution over the interval $[2,20]$. For the Gaussian waveform, $\tau$ is drawn from log-uniform distribution over $[0.001,0.01]$ seconds. The amplitudes are chosen so that the maximum of the signal is one, and the signals are centered by setting $t_0=0.5$ s.

We partition the atoms into three subdictionaries, using the natural clustering by the morphology of the glitches rather than resorting to clustering by $k$-medoids.  The subdictionaries  are subsequently compressed with a low-rank matrix factorization. Since the sine Gaussian and the ring-down signals admit negative values while for the Gaussian glitches are nonnegative, in the low-rank approximation we require non-negativity and sparsity of the coefficient matrix $\mH^{(j)}$ for each class, but non-negativity of $\mW^{(j)}$ only for the feature vectors of the Gaussian class. 
The number of feature vectors in each low-rank  approximation is set for each subdictionary based on an estimation of its effective rank. More specifically, let $K$ be the upper bound for the number of feature vectors in each subdictionary, chosen, e.g., based on computational considerations. If $\sigma_k^{(j)}$ is the $k$th singular value of the $j$th subdictionary matrix $\mD^{(j)}$, the rank $k_j$ of low-rank approximation of $\mD^{(j)}$ is set as
 \[
 k_j = \min\left\{K, \min\left\{ k \mid \frac{\sigma^{(j)}_k}{\sigma^{(j)}_1} <\delta\right\} \right\}
 \]
 for some fixed $\delta$, $0<\delta<1$.  Here we use the values $\delta = 0.001$ and $K=50$. We remark that while computing the full SVD of the subdictionaries is an expensive task, estimating the first few singular values is not a demanding task and often the signals in each subdictionary are quite similar, thus only a few singular values are required. The dimensions of the reduced dictionaries in this case are $k_{\rm SG} = 50$, $k_{\rm G} = 11$ and $k_{\rm RD} = 50$.  Using the training set, we compute the dictionary reduction error mean and covariance.

To demonstrate that including the modeling error in the identification problem makes a significant difference in the performance of the task, which in this case is to identify the morphology of a given glitch using the reduced subdictionaries, we generate 20\,000 noisy signals by drawing randomly  the parameters of the generative model and add scaled Gaussian white noise noise with standard deviation $\sigma = 0.05\, h_0$,
after which we normalize the signals  so that their maximum value is $1$. Hence, each glitch is corrupted with slightly different levels of scaled white noise and, unlike in the previous example, the test signals are out-of-sample data.

 To classify $b$, we find the least squares solution of the linear systems 
\[
 b \approx \mW^{(j)} h^{(j)} = \widehat b^{(j)}, \; j\in\{{\rm SG},{\rm G}, {\rm RD}\},
\]
whitened to compensate for the DCE correction, using the IAS algorithm with positivity and sparsity promotion. We compute the approximations $\widehat b^{(j)}$ and, as in \cite{Llorens-Montequado}, compare them to $b$ using the structural dissimilarity index proposed in \cite{Wang},
\[
 {\rm DSSIM}(b,\widehat b^{(j)}) = \frac 12 \big(1 - {\rm SSIM}(b,\widehat b^{(j)})\big).
 \]
The datum $b$ is assigned to the class with smallest dissimilarity index. 

To acknowledge the fact that the noise level is not known exactly, we overestimate it slightly, using 
\begin{equation}
	\tilde{\sigma} = 1.05 \sigma \, .
\end{equation}
Below, the confusion matrices  with noise levels $\sigma =0.05\,h_0$, $\sigma = 0.02\,h_0$ and $\sigma 0.01\,h_0$ are given, the rows corresponding to the classification results, and columns to the true classes:
\begin{center}
	\begin{tabular}{c|ccc}
		 	&SG	&G	&RD \\
		\hline
		SG	&6775	&0	&4 \\
		G	&57		&6535	&0 \\
		RD	&129	&1	&6499
	\end{tabular} \qquad
	\begin{tabular}{c|ccc}
			&SG	&G	&RD \\
		\hline
		SG	&6749	&0	&2 \\
		G	&0		&6598	&0 \\
		RD	&37		&0	&6499
	\end{tabular} \qquad
		\begin{tabular}{c|ccc}
		 	&SG	&G	&RD \\
		\hline
		SG	&6755	&0	&19 \\
		G	&0		&6599	&0 \\
		RD	&29		&0	&6598
	\end{tabular}
\end{center}
In particular, the success rates are 99.0\%, 99.2\%, and 99.8\%, respectively. 
 
Interestingly, the compensation of the DCE in this example is crucial for the success of the algorithm. In fact, when omitting the  DCE correction the classification goes egregiously astray even for $\sigma=0.01$, as illustrated by the following confusion matrix  
  \[
 \begin{array}{c|ccc}
  &{\rm SG} & {\rm G} & {\rm RD} \\
  \hline
  {\rm SG} &2 & 6797 & 3\\
  {\rm G} & 6583  &0 & 0 \\
  {\rm RD} & 6071 & 337 & 207
  \end{array}.
  \]
  
\subsection{Group sparsity promotion for hyperspectral data classification} 

In this computed experiment we illustrate how to use the group structural prior to filter out from a partitioned dictionary those subdictionaries that play no role in the dictionary representation of a new datum.  The dictionary is the Pavia University dataset, a hyperspectral image dataset collected in 2002. A sensor, known as the Reflective Optics System Imaging Spectrometer (ROSIS-3) \cite{ROSIS}, during a flight over the city of Pavia, Italy, collected 115 spectral bands covering wavelengths from 430 to 860 nm at locations above the University of Pavia.  Of the spectral bands, 12 of them corresponding to noisy channels were removed, leaving 103 spectral bands per each pixel.  The image area consists of $610 \times 340$  pixels with spatial resolution of 1.3 m. Of all the hyperspectral samples corresponding to the pixels, 40,776 of them, or approximately 20\%, were labeled according to the ground truth into 9 urban classes,   including asphalt, meadows, gravel, trees, metal sheet, bare soil, bitumen, brick, and shadow. 
 
\begin{figure}
\centerline{
\includegraphics[width = 16cm]{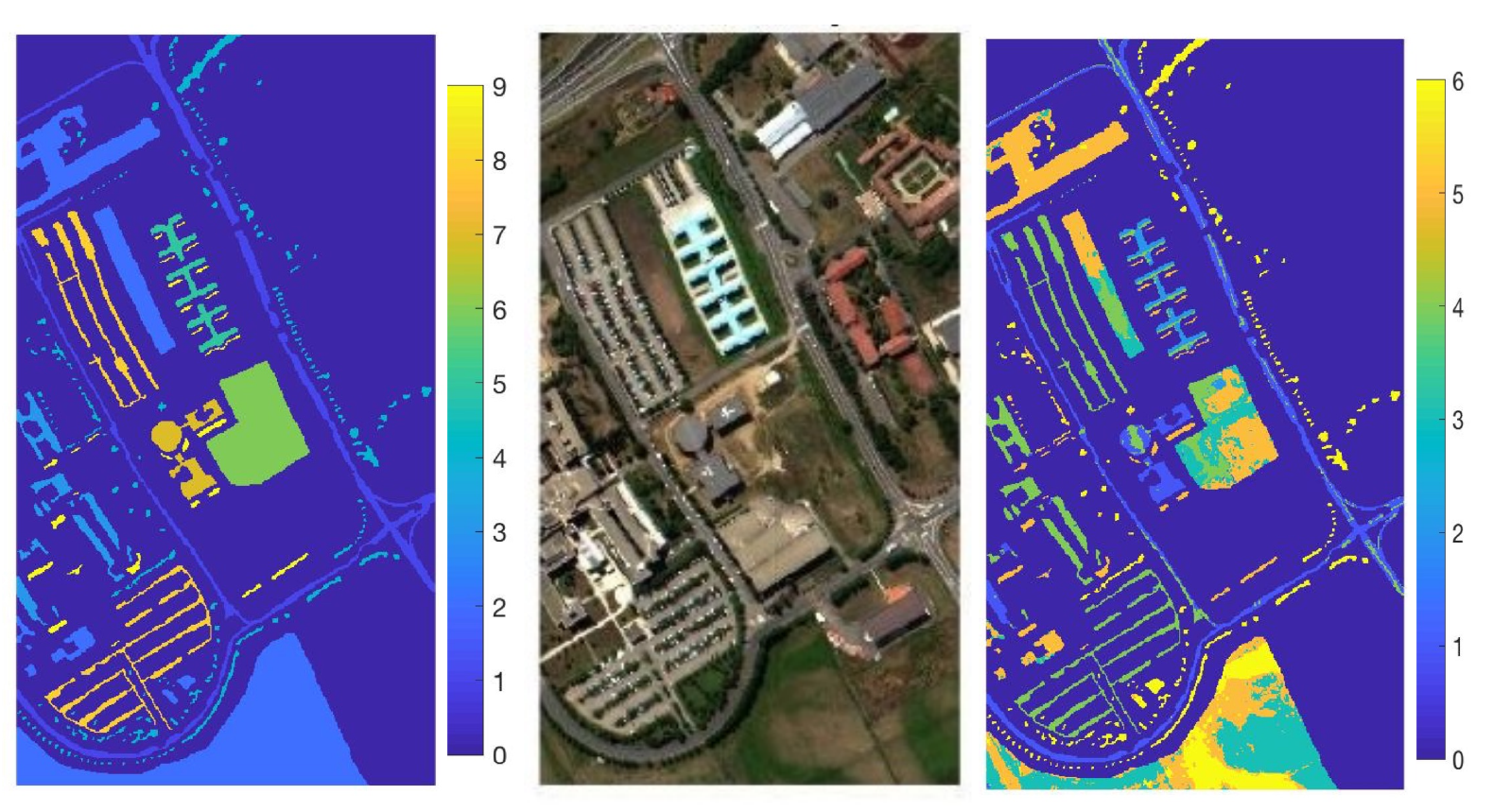}
}
\caption{\label{fig:Pavia} The partitioning of the labeled pixels of the University of Pavia areal view according to ground truth labeling (left) and unsupervised clustering (right). The ground truth classes correspond to Asphalt (1), Meadows (2), Gravel (3), Trees (4), Metal Sheet (5), Bare Soil (6), Bitumen (7), Brick (8), Shadows (9). Pixels in class 0 are those for which no ground truth based label is available.}
\end{figure}

While testing whether the hyperspectral data has enough resolution to resolve the ground truth classes by comparing the annotation with the results of unsupervised clustering with different number of classes, it was observed that several of the 9 urban classes corresponding to the ground truth were confounded, suggesting that effectively it would be more reasonable to reduce the partitioning to 6 classes, not necessarily in line with the ground truth. The ground truth clustering into 9 classes, together with the pseudoimage of the partitioning into 6 classes obtained by unsupervised clustering ($k$-medoids with $\ell^1$-distance matrix) are displayed in the left and right panels of Figure~\ref{fig:Pavia}. A comparison of the two pseudocolor images with an RGB image of the same area indicates that the partitioning via unsupervised clustering is closer to the RGB image than the expert annotation. 
 
We select half of the atoms in each of the six subdictionary for training, reserving the remaining for testing the effect of structural cone condition on the classification of the remaining data. Denoting the $j$th training subdictionary by $\mD^{(j)}$,  we compute its rank-$k$ PCA approximation,
\[
 \mD^{(j)} \approx \mU^{(j)}_k \mSigma_k^{(j)} \big(\mV^{(j)}_k\big)^\mT = \mW^{(j)} \mH^{(j)},
\]
where the truncation index $k$ is chosen based on the singular values $\sigma^{(j)}_\ell$ such that   
\[
 \sigma_k^{(j)} \geq \delta \sigma_1^{(j)} > \sigma_{k+1}^{(j)}, \quad \delta = 5\times 10^{-3}.
 \]
By construction, the  SVD of the coefficient matrix $\mH^{(j)}$ is simply given by
\[
 \mH^{(j)} = \mI_k \mSigma_k^{(j)} \left(\mV^{(j)}_k\right)^\mT,
\] 
therefore the cone condition defines the structural prior covariance corresponding to the $j$th class as
\[
 \mC^{(j)} = {\rm diag}\left((1,\sigma^{(j)}_2/\sigma^{(j)}_1, \ldots, \sigma^{(j)}_k/\sigma^{(j)}_1,0,\ldots, 0)\right) + \epsilon \mI,
\]

To test the group promotion power of the structural cone prior, we randomly select 100 hyperspectra from the testing data from each subdictionary and solve the compressed dictionary coding problem using the group sparsity promoting IAS with and without the structural cone prior. We use the value of the hyperparameter $\theta_j$, $1\leq j\leq 6$ as a proxy for the relevance of the $j$th class, scaling the vector $\theta$ of each draw by its largest component, so as have to have infinity norm 1, and look at how many classes have large enough $\theta$ to be deemed as relevant. 

\begin{figure}[ht!]
\centerline{
\includegraphics[width =  19cm]{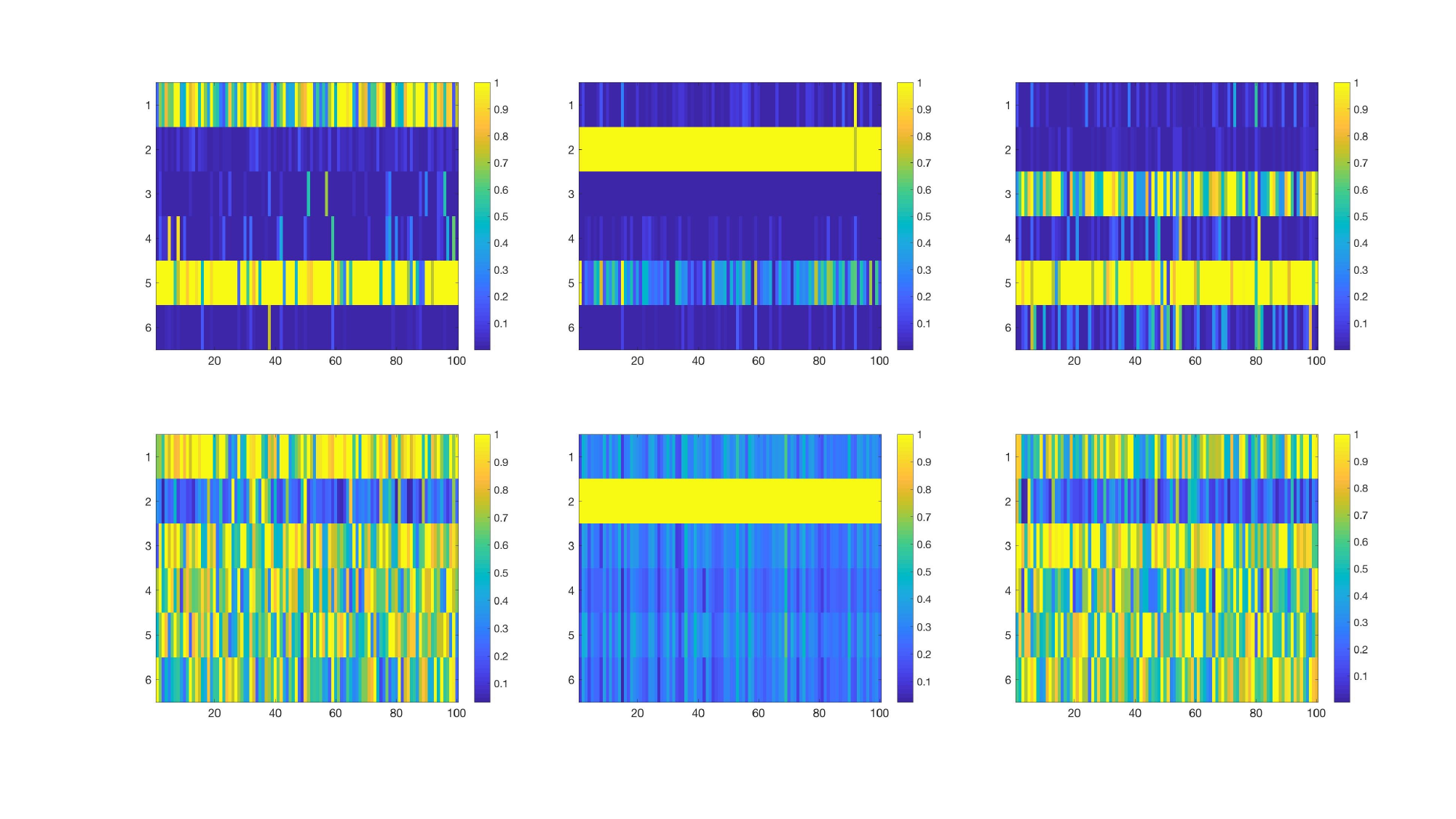}
}
\caption{\label{Class1-3} Each column shows the  hyperparameters for the six classes scaled to have infinity unit norm. The panels in each columns correspond to data from subdictionaries 1, 2 and 3, respectively. The top rows reports the result for IAS with the structural code prior, while the results in the bottom row are obtained with IAS without structural code prior.}
 \end{figure}

\begin{figure}[ht!]
\centerline{
\includegraphics[width =  19cm]{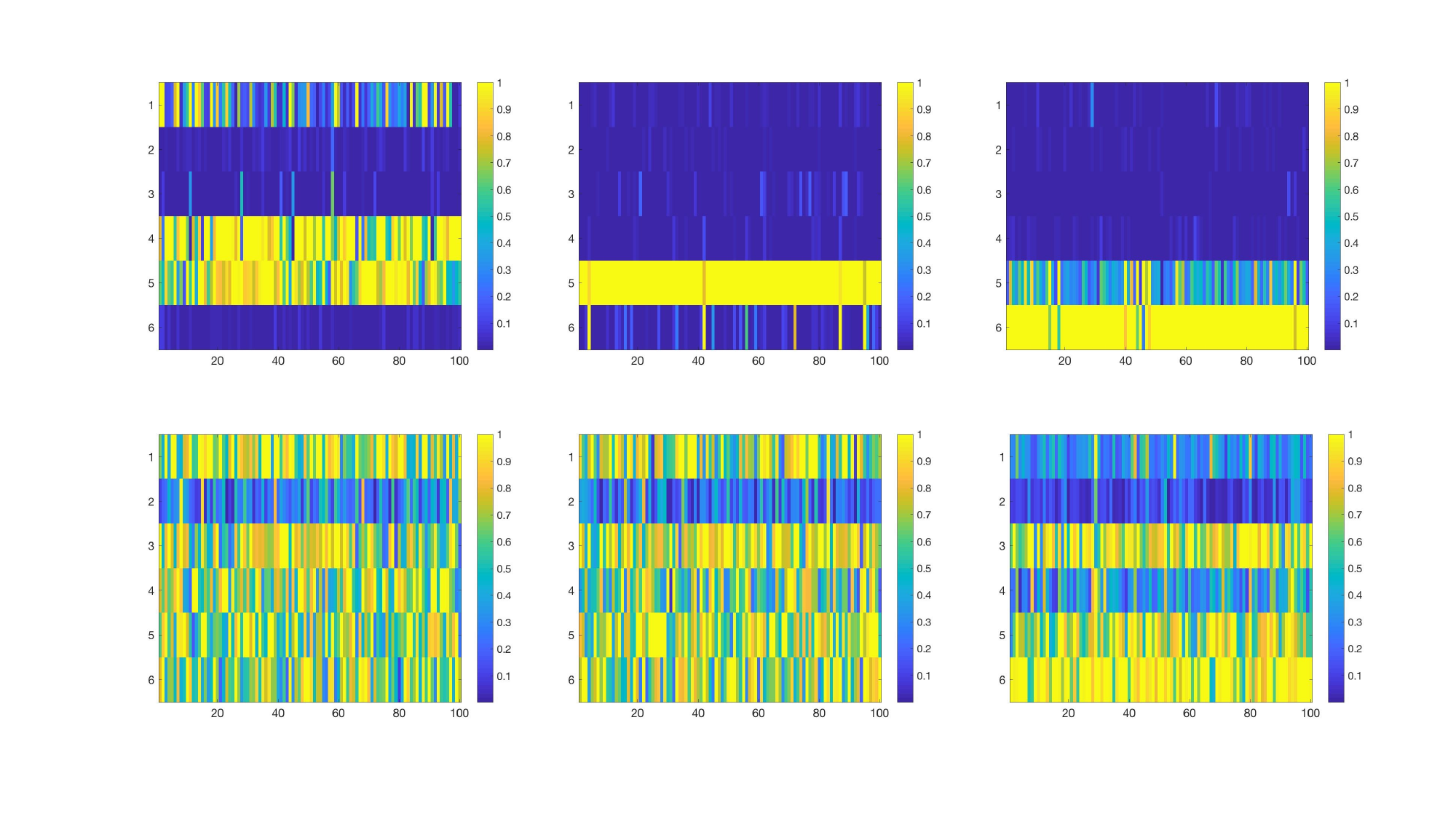}
}
\caption{\label{Class4-6} Each column shows the hyperparameters for the six classes, scaled to have infinity unit norm. The panels in each columns correspond to data from subdictionaries 4, 5 and 6, respectively. The top rows reports the result for IAS with the structural code prior, while the results in the bottom row are obtained with IAS without structural code prior.}
\end{figure}

Figure~\ref{Class1-3} shows the normalized vectors $\theta$  for the six classes computed by the IAS algorithm with the structural cone prior (top row) and without (bottom row) when the 100 data vectors are random draws of atoms from the testing set subdictionaries 1-3. The analogous results for subdictionaries 4-6 are shown in Figure~\ref{Class4-6}. The yellower the segment corresponding to the class, the higher the probability that the class plays a significant role, while the bluer the segment, the more likely the corresponding class is irrelevant for the datum on hand.

\begin{figure}[ht!]
\centerline{
\includegraphics[width =  16cm]{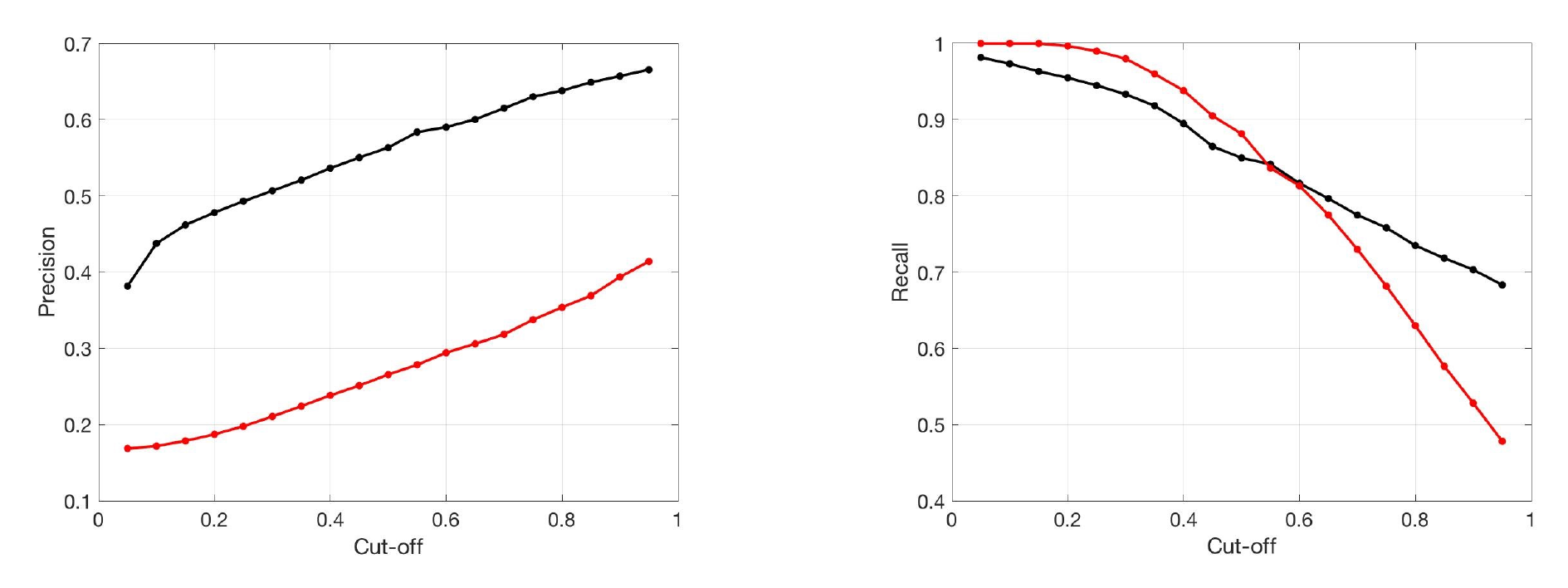}
}
\caption{\label{fig:Prec Recall} Precision and recall of the two group sparsity algorithms as a function of the cut-off value $\theta^*$. The black curves correspond to the group sparsity when the structural prior based on the cone condition is used, while the red ones correspond to the case when the structural prior covariances are replaced by an identity. We observe that including the cone condition stringly improves the discarding of irrelevant classes (higher precision), without significantly deteriorating the recall.}
\end{figure}

A popular way to quantify the accuracy of a classification algorithm are the {\em precision} or {\em positive predictive value}, and {\em precision}  or {\em sensitivity}, defined as
\[
{\rm Precision} = \frac{\mbox{\# relevant retrieved classes}}
{\mbox{\# retrieved classes}}, \quad
{\rm Recall} = \frac{\mbox{\# relevant retrieved classes}}
{\mbox{\# relevant classes}}.
\]
Hence, the precision indicates how effectively the algorithm deflates the dictionary without missing the correct subdictionary, while recall indicates how often the algorithm misses the true subdictionary, regardless of its ability to deflate. Ideally, both indices should be close to one. Figure~\ref{fig:Prec Recall} shows the precision and recall of the group sparsity algorithm with and without the structural cone condition, as a function of the cut-off value $\theta^*$. The plots show that with a typical value $\theta^* = 0.3$, the algorithm with the cone condition reduces the problem to approximately two subdictionaries (precision $\approx 0.5$) while without the cone condition, roughly 5 subdictionaries are retrieved (precision $\approx 0.2$.) On the other hand, the algorithm without the cone condition is likely to retrieve the true class (recall $\approx 1$) as it does a poor job in excluding classes, while the algorithm with the cone condition discards the correct class in roughly one case out of 20 (recall $\approx 0.95$.)

\begin{figure}[ht!]
\centerline{
\includegraphics[width =\textwidth]{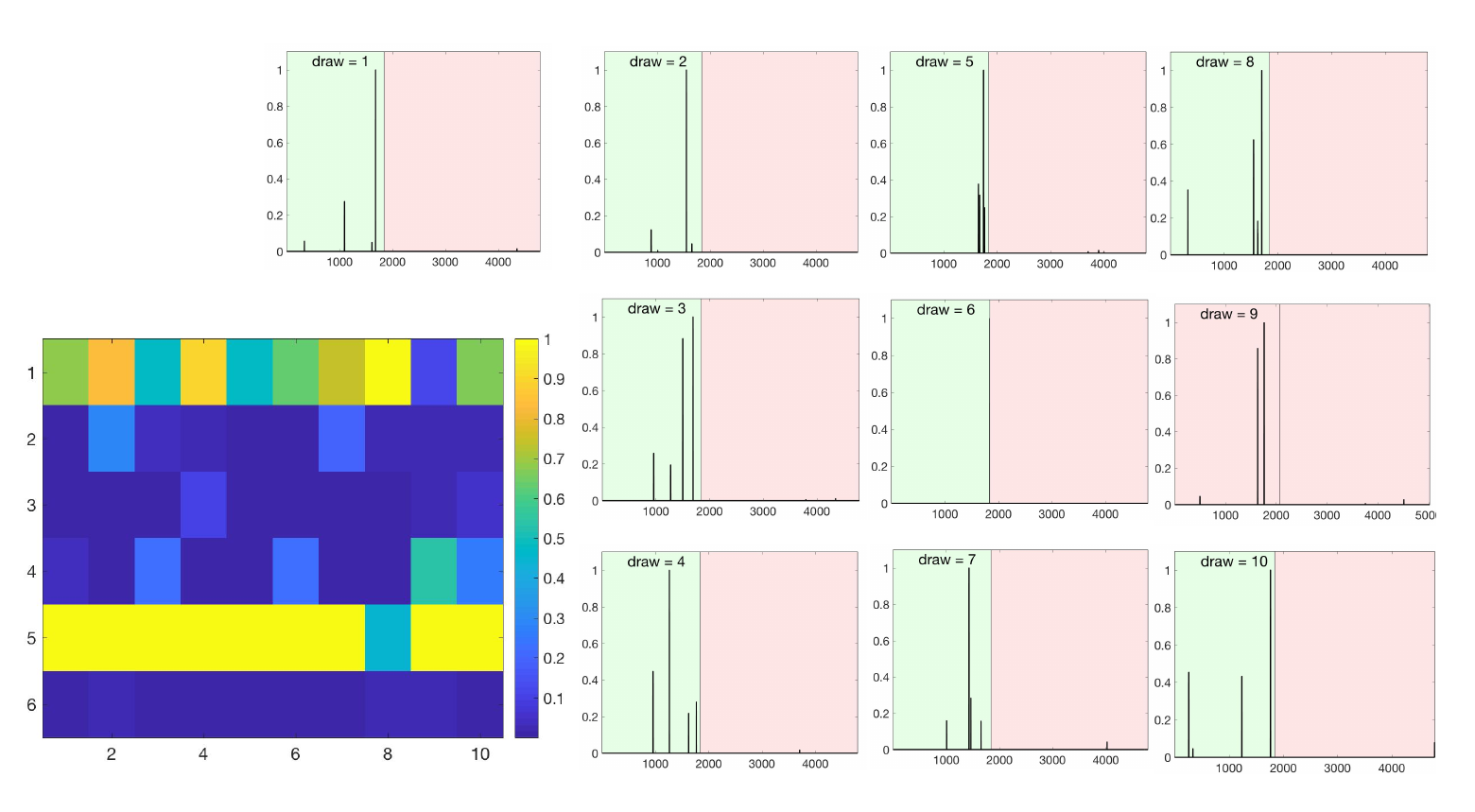}
}
\caption{\label{fig:deflated} Ten randomly drawn hyperspectra from the class 1 test subdictionary. The result of the relevant class identification shows that the first class is routinely confused with the fifth class. Nonetheless, when running the dictionary identification test using the hybrid IAS, the algorithm systematically proposes non-zero coefficients from the correct class (green background) rather than from the incorrect one (red background), except in the case of draw number 9 when the correct class was classified as irrelevant and discarded. The cut-off value for relevant classes in this example is $\theta^* = 0.3$.
}
\end{figure}

To complete this example, we run the full dictionary identification problem as follows: First, we draw randomly ten hyperspectra from the first test dictionary sample, and reduce the dictionary by using the group sparsity algorithm including the structural prior covariance. After discarding the subdictionaries with $\theta_j/\theta_{\rm max} <0.3$, we run the deflated dictionary identification algorithm using the hybrid IAS algorithm combining the gamma and inverse gamma hyperpriors to guarantee a strong sparsification of the coefficient vectors. In Figure~\ref{fig:deflated}, we show both the scaled group sparsity vectors as well as the outcomes of the deflated dictionary identification problems. In the plot, the coefficients corresponding to the correct subdictionary are plotted on a green background, while the coefficients in an incorrect subdictionary are on a red background. Interestingly, when the correct subdictionary is identified by the group sparsity phase as relevant for explaining the datum, the deflated solver seems to recognize the correct class: The non-zero coefficients are primarily chosen to correspond to the correct class rather than the incorrect ones even when the group sparsity phase indicates a stronger preference for a wrong class.  The only case when coefficients from an incorrect class are proposed is when the correct class was deemed irrelevant.

\section{Conclusions}

Bayesian methods provide a versatile and rich environment for computational inverse problems, making it possible to integrate prior information about the unknowns, and to compensate for shortcomings in the models. Interpreting data science problems such as dictionary learning and dictionary matching as inverse problems opens new ways to introduce well developed Bayesian methods into this active and rapidly growing field. This article demonstrates how the Bayesian modeling error approach and structural prior information can be translated into effective tools to improve classification in the sense of subdictionary identification, and to solve efficiently large scale dictionary matching problems with sparsity promoting priors. The Bayesian approach opens also a way of addressing an important  and novel area of research, the uncertainty quantification in data science, crucial for the reliability of data-driven methods in areas like medical imaging and structural health research where the assessment of the reliability is of central importance.

\section*{Acknowledgements}

This work made use of the High Performance Computing Resource in the Core Facility for Advanced Research Computing at Case Western Reserve University. The work of DC was supported in part by NSF Award DMS 1951446. The work of ES was supported in part by NSF Award DMS 2204618.

\end{document}